\documentclass[11pt]{article}
\usepackage{algorithm,algorithmicx,algpseudocode,amsmath,amssymb,amsthm}
\usepackage{bm,color,dsfont,epsfig,fullpage,graphicx,pifont}
\usepackage{makecell}
\usepackage[toc,page]{appendix}
\usepackage{url}
\usepackage{multirow}
\usepackage{comment}
\usepackage{enumitem}
\usepackage{xcolor}
\usepackage{epstopdf}
\usepackage{subfig}
\usepackage[noabbrev]{cleveref}
\crefname{equation}{}{}

\allowdisplaybreaks

\newtheorem{lemma}{Lemma}
\newtheorem{proposition}{Proposition}
\newtheorem{theorem}{Theorem}

\def\lb{\left(}
\def\rb{\right)}

\def\ln{\left\|}
\def\rn{\right\|}

\def\P{\mathcal{P}}

\def\T{\mathcal{T}}

\def\Si{\Sigma}

\DeclareMathOperator{\diag}{diag}

\DeclareMathOperator{\supp}{supp}
\begin{document}

\title{Robust Multi-Dimensional Scaling via Accelerated Alternating Projections}
\author{Tong Deng \thanks{T. Deng and T. Wang are with the School of Mathematics, Southwestern University of Finance and Economics, Chengdu, Sichuan, China.} \and Tianming Wang \footnotemark[1]}

\maketitle

\begin{abstract}
We consider the robust multi-dimensional scaling (RMDS) problem in this paper. The goal is to localize point locations from pairwise distances that may be corrupted by outliers. Inspired by classic MDS theories, and nonconvex works for the robust principal component analysis (RPCA) problem, we propose an alternating projection based algorithm that is further accelerated by the tangent space projection technique. For the proposed algorithm, if the outliers are sparse enough, we can establish linear convergence of the reconstructed points to the original points after centering and rotation alignment. Numerical experiments verify the state-of-the-art performances of the proposed algorithm.  
\end{abstract}

\section{Introduction}

Multi-dimensional scaling (MDS) refers to the localization of points from their pairwise distances, and it has applications in wireless communication, chemistry, computer graphics, and so on \cite{Dokmanic2015}. The properties of classic MDS have been well studied.

In reality, the measured pairwise distances may be incomplete, noisy, or, due to  sensor malfunction, corrupted by outliers. In this paper, we consider the robust reconstruction of the point locations from distances corrupted by outliers, i.e., the robust multi-dimensional scaling (RMDS) problem. As shown in \cite{cayton2006robust}, even a single outlier in the measured distances will cause the false information to spread to all the reconstructed point locations in the MDS calculation. 

Suppose that we have a set of points $\{x_i\}_{i=1}^n\subseteq \mathbb{R}^{r}$. Denote the squared Euclidean distance matrix (EDM) as $D^{\star}$. The $(i,j)$-th entry of $D^{\star}$ is equal to $\|x_i-x_j\|_2^2$. If the pairwise distances are corrupted by outliers, the observed EDM $D$ is equal to $D^{\star}+S^{\star}$ for some outlier matrix $S^{\star}$. One can see that the considered problem may be solved by methods for robust principal component analysis (RPCA) \cite{Candes2011,netrapalli2014non,Yi2016,accaltprj2019}. However, naively adopting a RPCA solver for the RMDS problem is at the risk of neglecting the inner structure of the EDM, and may result in degraded reconstruction performances \cite{Li2017}. 

To combat outliers in the EDM, \cite{Forero2012} casts the problem as an unconstrained optimization, where the variables are the data matrix of size $n\times r$, and the outlier matrix of size $n\times n$. The data fidelity term of the objective function is the squared Frobenius norm of the difference between the observed distance matrix, and the distance matrix formed by the data matrix plus the outlier matrix. The regularization term of the objective function is the $l_1$ penalty for the outlier matrix. The optimization is conducted using the majorization-minimization approach. Later, the iterative scheme of \cite{Forero2012} is improved by \cite{Mandanas2017}, replacing the squared Frobenius norm with the more outlier-robust M-estimators. A more recent work \cite{Kong2019}, formulates the RMDS problem as a constrained optimization. Besides the outlier matrix, another sought-after matrix is constrained in a way such that it is the distance matrix formed by $n$ points with dimension less than or equal to $r$.  Compared to \cite{Forero2012,Mandanas2017}, \cite{Kong2019} shows improved performances. Weights and bounds for the entries of the distance matrix can also be handled easily in \cite{Kong2019}. Different from the optimization perspectives of the aforementioned works, \cite{Blouvshtein2019} proposes to first detect the outliers by the broken triangular inequalities among the distances, and then compute a weighted MDS without the detected corrupted distances. However, as shown in \cite{Kong2019}, oftentimes such an approach is not enough to yield satisfactory reconstruction performances. 

Inspired by classic MDS theories, and state-of-the-art nonconvex solvers \cite{netrapalli2014non,accaltprj2019} for the RPCA problem, we propose a nonconvex algorithm that alternates between finding the outlier matrix, and the Gram matrix that generates the outlier-free EDM. Our contributions can be summarized as the following.

\begin{itemize}
    \item For the proposed algorithm, we can establish linear convergence of the reconstructed points to the original points after centering and rotation alignment, if the outliers are sparse enough. The relationship between the amount of tolerable outliers and the properties of the original points, is also made clear in the theoretical guarantees.
    \item We numerically verify the performances of the proposed algorithm in the considered outlier only setting, and demonstrate its advantages compared to other state-of-the-art solvers in the noise plus outlier setting.
\end{itemize}

\subsection{Problem Formulation and Assumptions}

Some necessary notations are first introduced. We then describe our problem formulation, and the assumptions to derive the theoretical guarantees.

\noindent\textbf{Notations.} The set of symmetric matrices of size $n$ is denoted by $\mathbb{S}^{n\times n}$. The set of rotation matrices of size $r$ is denoted by $\mathcal{O}(r)$, i.e., $\mathcal{O}(r)=\{G\in\mathbb{R}^{r\times r}~|~G^TG=I_r\}$. $\forall Z\in\mathbb{S}^{n\times n}$, $\supp(Z)$ is the indices of the nonzeros in $Z$, and $\diag(Z)\in\mathbb{R}^n$ is the vector that contains the diagonal entries of $Z$. $\bm{1}\in\mathbb{R}^n$ denotes the all-one vector, and $J=I_n-\frac1n\bm{1}\bm{1}^T$. $\forall Z\in\mathbb{R}^{n\times r}$, $\|Z\|_{2,\infty} := \max_{1\leq i\leq n} \|e_i^TZ\|_2.$ For any matrix $M$, the spectral norm, the Frobenius norm, and the entrywise infinity norm of $M$ are denoted by $\| M \|_2$, $\| M \|_{\mathrm{F}}$, and $\|M\|_{\infty}$, respectively.

Recall that the set of points is $\{x_i\}_{i=1}^n\subseteq \mathbb{R}^{r}$, and the corresponding EDM is $D^{\star}$. As it turns out, it is possible to recover another set of points centered at zero and preserve the pairwise distances of $\{x_i\}_{i=1}^n$. Denote the data matrix as $X\in\mathbb{R}^{n\times r}$, whose $i$-row is $x_i^T$. Let $c=\frac1n \sum_{i=1}^n x_i$, and then denote the centered data matrix as $X_c\in\mathbb{R}^{n\times r}$, whose $i$-row is $(x_i-c)^T$. Define the operator $\mathcal{A}:\mathbb{S}^{n\times n}\rightarrow\mathbb{S}^{n\times n}$ such that
\begin{equation}\label{eq:A}
\mathcal{A}(Z) = \text{diag}(Z)\bm{1}^T+\bm{1}\text{diag}(Z)^T-2Z,\quad\forall Z\in\mathbb{S}^{n\times n}. 
\end{equation}
One can verify that $D^{\star}=\mathcal{A}(XX^T)=\mathcal{A}(X_cX_c^T)$. Hence, in many applications, it often suffices to reconstruct $X_c$ from $D^{\star}$.

Denote $L^{\star}:=X_cX_c^T$, and define the operator $\mathcal{B}:\mathbb{S}^{n\times n}\rightarrow\mathbb{S}^{n\times n}$ such that
\begin{equation}\label{eq:B}
\mathcal{B}(Z) = -\frac12\cdot J Z J,\quad\forall Z\in\mathbb{S}^{n\times n}.  
\end{equation}
Since $L^{\star}\bm{1}=0$, by Lemma \ref{lem:A}, $\mathcal{B}(D^{\star}) = \mathcal{B}(\mathcal{A}(L^{\star})) = L^{\star}$. Furthermore, $X_c$ can be reconstructed, after rotation alignment, from the eigen-decomposition of $L^{\star}$. Without loss of generality, we assume that $X_c$ is of full column rank, then $L^{\star}$ is a rank-$r$ positive semi-definite matrix. Denote the eigen-decomposition of $L^{\star}$ as 
$U^{\star}\Lambda^{\star}(U^{\star})^T$, where $U^{\star}\in\mathbb{R}^{n\times r}$, $\Lambda^{\star}=\text{diag}(\lambda_1,\cdots,\lambda_r)$, and $\lambda_1^{\star}\geq \cdots\geq \lambda_r^{\star}>0$. Then denote $X^{\star}=U^{\star}(\Lambda^{\star})^{\frac12}$. One can show that there exists a rotation matrix $Q^{\star}\in\mathcal{O}(r)$ such that $X_c = X^{\star}Q^{\star}$.

In the considered setting, the observed EDM $D$ is equal to $D^{\star}+S^{\star}$ for some outlier matrix $S^{\star}\in\mathbb{S}^{n\times n}$. Our aim is to recover $L^{\star}$ (eventually, $X_c$) from $D=\mathcal{A}(L^{\star})+S^{\star}$. To derive our theoretical guarantees, similar to the incoherence and sparsity assumptions commonly made in the RPCA literature \cite{netrapalli2014non,accaltprj2019}, we assume the following about $L^{\star}$ and $S^{\star}$. 

\paragraph{Assumption~1.} Suppose $L^{\star}$ has the eigen-decomposition $U^{\star}\Lambda^{\star}(U^{\star})^T$, where $U^{\star}\in\mathbb{R}^{n\times r}$, $\Lambda^{\star}=\text{diag}(\lambda_1,\cdots,\lambda_r)$, and $\lambda_1^{\star}\geq \cdots\lambda_r^{\star}>0$. We assume that $L^{\star}$ is $\mu$-incoherent\footnote{One can show that $\mu\in[1,\frac{n}{r}]$. Typically, $\mu=O(1)$.}, i.e.,
$$
\|U^{\star}\|_{2,\infty}\leq \sqrt{\frac{\mu r}{n}}.
$$
From this assumption, one can immediately get
$$
\|L^{\star}\|_{\infty}=\max_{i,j}\|e_i^TU^{\star}\Lambda^{\star}(U^{\star})^Te_j\|\leq \frac{\mu r}{n}\lambda_1^{\star}.
$$

\paragraph{Assumption~2.} The outlier matrix $S^{\star}\in\mathbb{S}^{n\times n}$ is $\alpha$-sparse, i.e., it has no more than $\alpha n$ nonzero entries per row (and column).

\section{Algorithm}

Our proposed algorithm, described in Algorithm \ref{alg:RMDS-AAP}, is inspired by classic MDS theories, and state-of-the-art nonconvex solvers \cite{netrapalli2014non,accaltprj2019} for RPCA.

\begin{algorithm}[!ht]
\caption{RMDS via Accelerated Alternating Projections (RMDS-AAP)} \label{alg:RMDS-AAP}
\begin{algorithmic}[1]
\State \textbf{Inputs:} EDM $D$, target rank $r$, threshold parameter $\xi^0>0$, and decay rate $\gamma\in(0,1)$.
\State \textbf{Initialization:} $S^{0}=\T_{\xi^{0}}(D)$, $L^{1}=\mathcal{H}_r^{+}\mathcal{B}(D-S^0)$.
\For{$k=1,2,\cdots$}
    \State $S^{k} = \T_{\xi^{k}}(D-\mathcal{A}(L^k))$, where $\xi^{k} = \xi^0\cdot\gamma^{k}$
    \State $L^{k+1}=\mathcal{H}_r^{+}\P_{T^k}\mathcal{B}(D-S^k)$
\EndFor
\end{algorithmic}
\end{algorithm}

At initialization, with the hard thresholding function $\mathcal{T}_{\xi}(z)~(\xi>0):\mathbb{R}\rightarrow\mathbb{R}$ defined as
$$
\mathcal{T}_{\xi}(z) = \left\{\begin{array}{cc}
z & |z|>\xi \\
0 & |z|\leq \xi
\end{array}\right.,
$$
large entries corrupted by outliers in $D$ are picked out. Then in the spirit of MDS, $L^1$ is computed from $D-S^0$. Here, the operator $\mathcal{B}$ is defined as in \eqref{eq:B}, and $\forall Z\in\mathbb{S}^{n\times n}$ with the eigen-decomposition $U\Lambda U^T$, where $U=[u_1,\cdots,u_n]\in\mathbb{R}^{n\times n}$, $\Lambda=\diag(\lambda_1,\cdots,\lambda_n)$, and $\lambda_1\geq \cdots\geq\lambda_n$,
$$
\mathcal{H}_r^{+}(Z) = \sum_{i=1}^r\max\{\lambda_i,0\}\cdot u_iu_i^T.
$$

For later iterations ($k\geq 1$), the threshold parameter $\xi^k$ is adjusted by the decay rate $\gamma$ to picked out the entries with outliers. Denote the eigen-decomposition of $L^k$ as $U^{k}\Lambda^k(U^k)^T$, where $U^k\in\mathbb{R}^{n\times r}$, the operator $\P_{T^k}$ defined as the following,
\begin{equation}\label{eq:projection}
\P_{T^k}(Z) := U^k(U^k)^TZ+ZU^k(U^k)^T-U^k(U^k)^TZU^k(U^k)^T,~\forall Z\in \mathbb{S}^{n\times n},
\end{equation}
is the projection onto the tangent space of the manifold of symmetric positive semi-definite matrices of rank $r$ at $L^k$. Such tangent space projection applied before the partial eigen-decomposition has been proven useful in deriving theoretical guarantees as well as reducing computation cost \cite{Wei2016,accaltprj2019,Cai2021}. 

For RMDS-AAP, the dominant cost is the computation of $L^k$ each iteration. At initialization, the flops are about $5n^2+O(n^2r)$, where the hidden constant comes from the partial eigen-decomposition. For later iterations, with the help of the tangent space projection, the partial eigen-decomposition of a $n\times n$ matrix is reduced to several matrix-matrix multiplications that costs about $5n^2+2n^2r+8nr^2$ flops, a QR factorization of a $n\times r$ matrix, and a small eigen-decomposition of a $2r\times 2r$ matrix. The total cost is about $5n^2+2n^2r+O(nr^2)$ flops. For completeness, we include the implementation details in Appendix \ref{sec:implementation}. Compared to \cite{accaltprj2019}, the main difference comes from applying $\mathcal{B}$ before the tangent space projection. 

For the theoretical guarantees of RMDS-AAP, we have derived the following results. 

\begin{theorem}\label{thm:main} 
Suppose that RMDS-AAP is provided with $\xi^0$ that satisfies $\|D^{\star}\|_{\infty} \leq \xi^0 \leq 3\|D^{\star}\|_{\infty}$, and $\gamma\in[\frac{1}{3},1)$. Denote $\kappa:=\frac{\lambda_1^{\star}}{\lambda_r^{\star}}$. If
$$
\alpha\leq \frac{1}{1624}\cdot\frac{\gamma}{\mu r\kappa^2},
$$
then for $\forall k\geq 0$, $\text{supp}(S^k)\subseteq \text{supp}(S^{\star})$,
$$
\|S^k-S^{\star}\|_{\infty} \leq (4\|D^{\star}\|_{\infty})\gamma^k,
$$
and
$$
\|L^{k+1}-L^{\star}\|_{\infty} \leq \frac{\|D^{\star}\|_{\infty}}{4}\gamma^{k+1}.
$$
\end{theorem}

\noindent\textbf{Remark 1.} The constant in the bound for $\alpha$ can be further optimized. The established bound $O(\frac{1}{\mu r\kappa^2})$ is better than the one $O(\frac{1}{\mu r^2\kappa^3})$ in \cite{accaltprj2019} for the RPCA problem, and the one $O(\frac{1}{\mu^2r^2\kappa^2})$ in \cite{Cai2021} for the robust recovery of low-rank Hankel matrices, showing the merits of our proof. When $\kappa=O(1)$, the bound matches the optimal one $O(\frac{1}{\mu r})$ in \cite{netrapalli2014non} for the RPCA problem.   

\begin{proposition}\label{prop}
Assume the same conditions of Theorem \ref{thm:main}. For $\forall k\geq 0$, further denote $X^{k+1}=U^{k+1}(\Lambda^{k+1})^{\frac12}$, and $X^{\star}=U^{\star}(\Lambda^{\star})^{\frac12}$. Suppose the rank-$r$ SVD of $\left(X^{\star}\right)^TX^{k+1}$ is $Y^{k+1}\widetilde{\Si}^{k+1} (Z^{k+1})^T$. Set $R^{k+1}=Y^{k+1}(Z^{k+1})^T$. Then
$$
\|X^{k+1}-X^{\star}R^{k+1}\|_{2,\infty} \leq \sqrt{\frac{\mu r\kappa\lambda_1^{\star}}{n}}\gamma^{k+1}.
$$
\end{proposition}

\noindent\textbf{Remark 2.} As mentioned in the problem formulation, $X_c=X^{\star}Q^{\star}$ for some rotation matrix $Q^{\star}$. Furthermore, one can show that the computed minimizer to
$$ 
\min_{G\in\mathcal{O}(r)}\|X^{k+1}-X_cG\|_F = \min_{G\in\mathcal{O}(r)}\|X^{k+1}-(X^{\star}Q^{\star})G\|_F 
$$
is $(Q^{\star})^TR^{k+1}$. Therefore, $X^{k+1}-X^{\star}R^{k+1}=X^{k+1}-X_c(Q^{\star})^TR^{k+1}$, and Proposition \ref{prop} actually establishes the linear convergence of the reconstructed points $X^{k+1}$ to $X_c$ after the best rotation alignment by $(Q^{\star})^TR^{k+1}$. Also, the contraction of error in the $l_2$ norm is uniform for all the points. 

\section{Numerical Experiments}

We perform tests on the plus sign example that appeared in \cite{Forero2012,Kong2019}. The tests are first conducted for the noiseless case, i.e., the distances are only corrupted by outliers, to verify our theoretical guarantees for RMDS-AAP. We then consider the more realistic case, i.e., the distances are corrupted by both noise and outliers, and show the empirical performances of RMDS-AAP.

\begin{figure}[!ht]
\subfloat{\includegraphics[width=.33\linewidth]{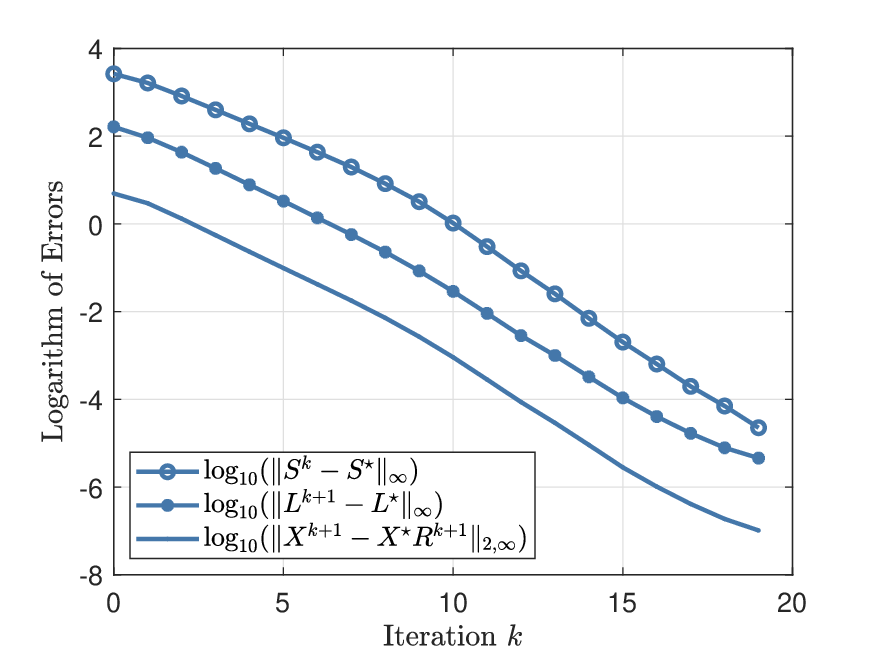}} \hfill
\subfloat{\includegraphics[width=.33\linewidth]{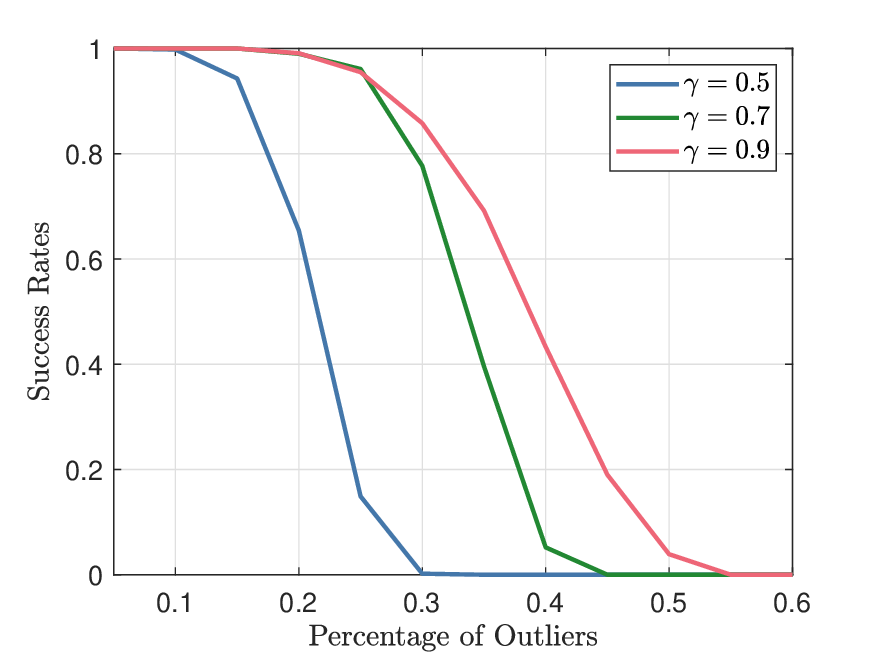}} \hfill
\subfloat{\includegraphics[width=.33\linewidth]{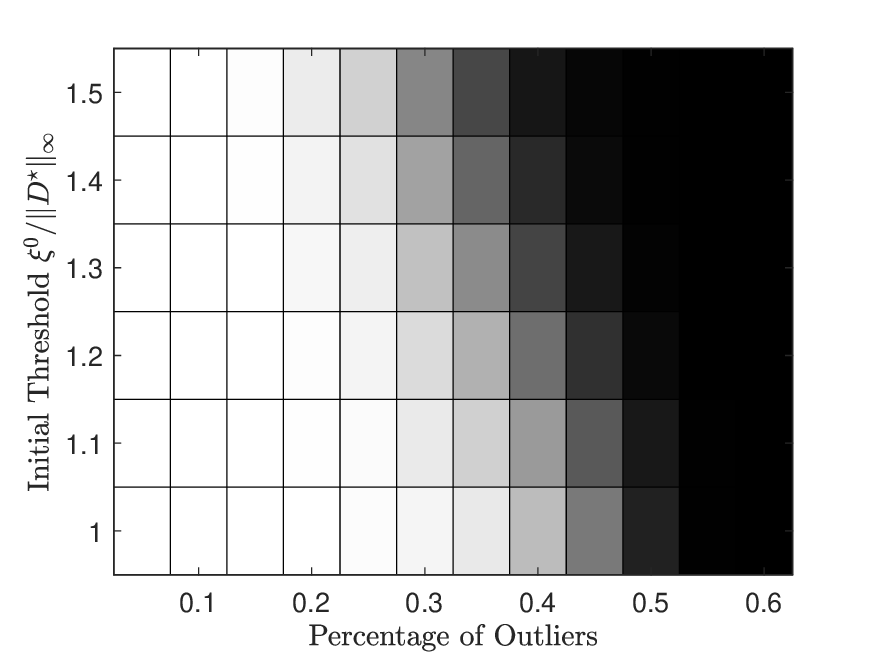}} \hfill
\caption{Performances of RMDS-AAP for the plus sign (101 points), where the distances are only corrupted by outliers.} \label{fig:noiseless}
\end{figure}

\noindent\textbf{Noiseless Case.} In this case, we consider a plus sign consists of $101$ 2D points $\{x_i\}_{i=1}^{101}$, which are centered at $c=[6~6]^T$, and have four end points at $[-19~6]^T$, $[31~6]^T$, $[6~-19]^T$, and $[6~31]^T$. The data matrix $X\in\mathbb{R}^{101\times 2}$ has $x_i^T$ as its $i$-th row, and $X_c = X-\bm{1}c^T$. The ground truth Gram matrix $L^{\star}=X_cX_c^T\in\mathbb{R}^{101\times 101}$ has incoherence parameter $\mu\approx 3$, and condition number $\kappa = 1$. To test the reconstruction performances of RMDS-AAP against outliers, $m$ out of the $N:=\frac{101\times 100}{2} = 5050$ distances
$
\left\{d_{ij} :=\|x_i-x_j\|_2~|~1\leq j<i\leq n\right\}
$
are randomly sampled, and each is added with an outlier whose value is uniformly selected within $[0~40]$. Denote the percentage of outliers as $p:=\frac{m}{N}$. For $p\in\{0.05,0.1,0.15,\cdots,0.6\}$, we test RMDS-AAP with different initial threshold values of $\xi^0$ and decay rate values of $\gamma$. The results, averaged among 1000 simulations, are reported in Fig.~\ref{fig:noiseless}. 

In the left subfigure, the logarithms of the averaged errors in terms of $\log_{10}(\|S^{k}-S^{\star}\|_{\infty})$, $\log_{10}(\|L^{k+1}-L^{\star}\|_{\infty})$, and $\log_{10}(\|X^{k+1}-X^{\star}R^{k+1}\|_{\infty})$ are plotted when $p=0.05$, $\xi^0 = 1.2\|D^{\star}\|_{\infty}$, and $\gamma=0.5$. One can see the linear convergence of the three terms, in agreement with Theorem \ref{thm:main} and Proposition \ref{prop}. In the middle and right subfigures, the reconstruction of each simulation is considered successful if, at convergence,
$$
\|X^{k+1}-X^{\star}R^{k+1}\|_{2,\infty} < 0.01\cdot\|X^{\star}\|_{2,\infty}.
$$
In the middle subfigure, $\xi^0$ is selected as $1.2\|D^{\star}\|_{\infty}$, and the success rates computed out of the 1000 simulations are compared for $\gamma\in\{0.5,0.7,0.9\}$. As predicted in Theorem \ref{thm:main}, one can see that larger $\gamma$ indeed admits successful reconstruction from more outliers. In the right subfigure, the success rates when $\gamma=0.9$, and $\xi^0\in\{\|D^{\star}\|_{\infty},1.1\|D^{\star}\|_{\infty},\cdots,1.5\|D^{\star}\|_{\infty}\}$ are shown, whereas the white color indicates success rate 1, and the black color indicates success rate 0. One can see that RMDS-AAP shows some desired robustness to the initial threshold value $\xi^0$.

\begin{figure}[!ht]
\subfloat{\includegraphics[width=.33\linewidth]{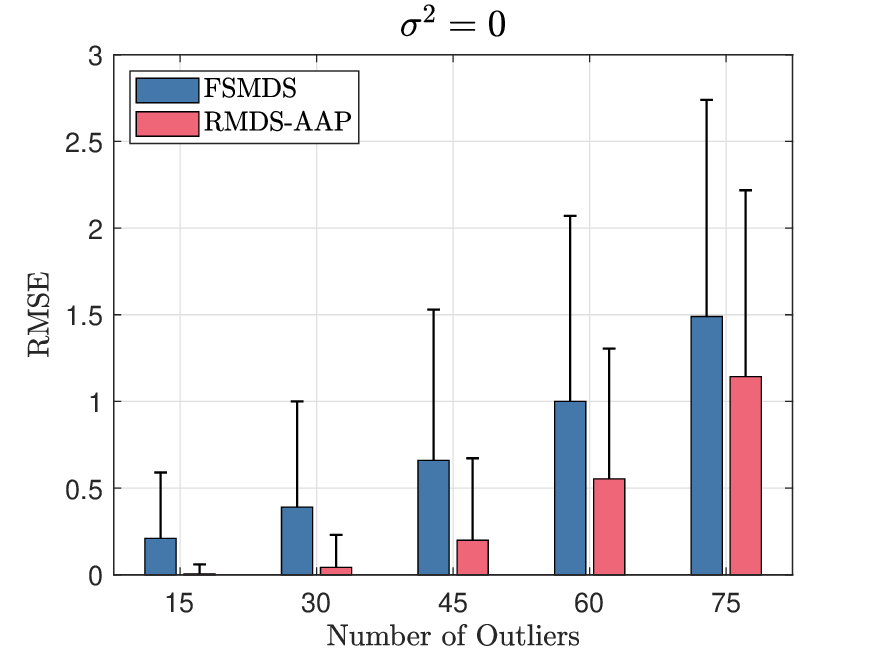}} \hfill
\subfloat{\includegraphics[width=.33\linewidth]{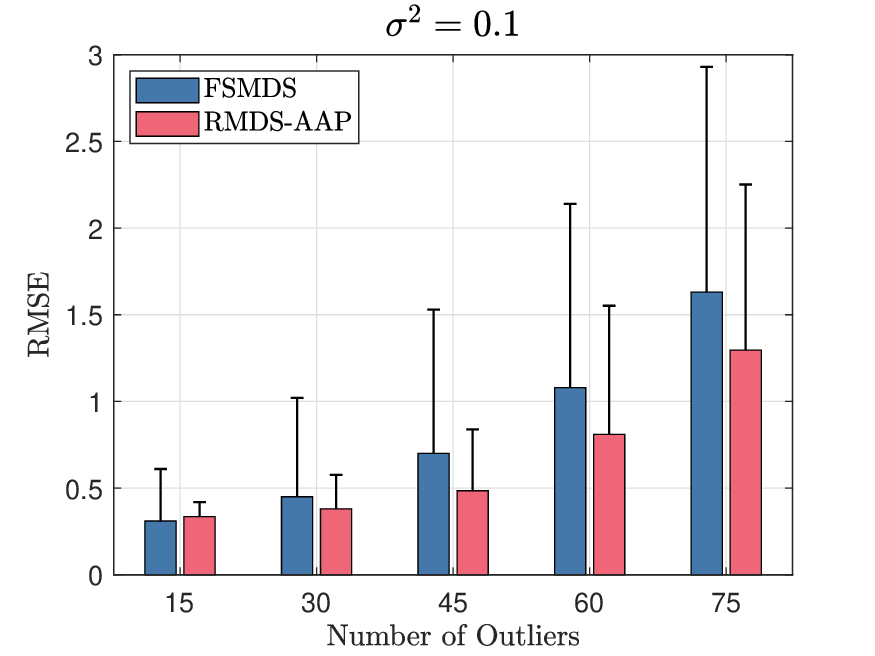}} \hfill
\subfloat{\includegraphics[width=.33\linewidth]{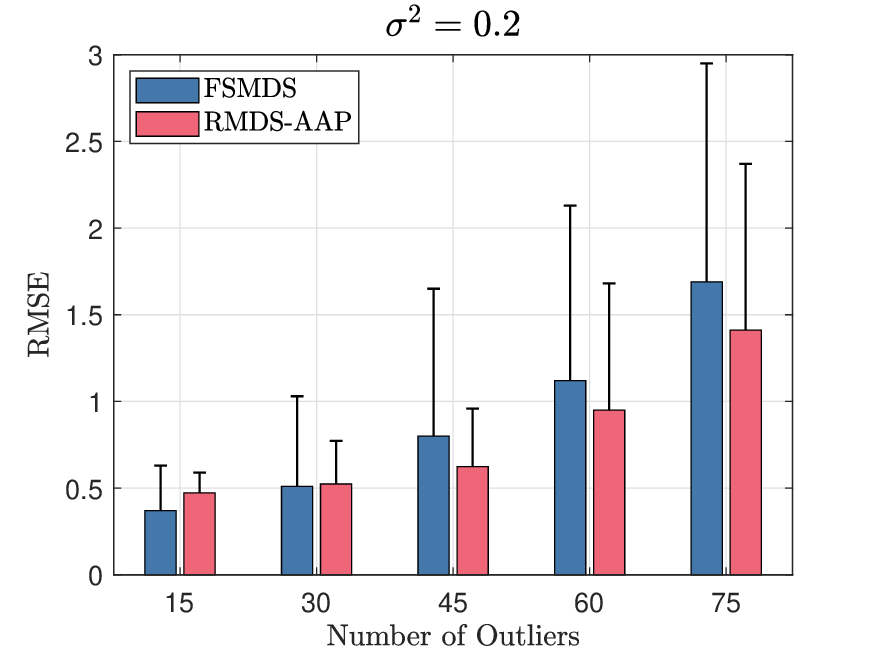}} \hfill
\caption{Performance comparisons for the plus sign (25 points) with 4 anchor points, where the noisy distances are further corrupted by outliers. The error bars show the standard deviation values of the two methods.} \label{fig:noisy}
\end{figure}

\noindent\textbf{Noisy Case.} We then empirically test the reconstruction ability of RMDS-AAP when the distances are also noisy. Following the setup of \cite{Kong2019}, 25 points centered at $c=[6~6]^T$ form the plus sign, with 4 anchor points at $[0~6]^T$, $[12~6]^T$, $[6~0]^T$, and $[6~12]^T$. For $1\leq j<i\leq n$, each $d_{ij}$ is first added with a zero-mean Gaussian noise with variation $\sigma^2\in\{0,0.1,0.2\}$. Then $m\in\{15,30,45,60,75\}$ out of the $N:=\frac{25\times 24}{2}-6=294$ distances, discounting for the 6 pairwise distances between the 4 anchor points, are sampled randomly to independently add with an outlier whose value is uniformly selected within $[0~20]$. We use $\xi^0=1.2\|D^{\star}\|_{\infty}$ and $\gamma=0.7$ for RMDS-AAP. At convergence, a linear mapping $\mathcal{T}$, consists of translation and rotation, is constructed to find the best alignment between the reconstructed 4 anchor points and the original 4 anchor points. Denote $\Omega$ as the indices of the 4 anchor points, and denote the $i$-th row of the $X^{k+1}$ at convergence as $(x_i^{\text{rec}})^T$. The error measure is computed at the other 21 points after alignment, i.e.,
$\text{RMSE} := \sqrt{\sum_{i\notin\Omega}\|\mathcal{T}(x_i^{\text{rec}})-x_i\|_2^2/21}.$ Since in this example, FSMDS \cite{Kong2019} shows the best performances, compared to RMDS \cite{Forero2012}, HQMMDS \cite{Mandanas2017}, and TMDS \cite{Blouvshtein2019}, we only compare RMDS-AAP with FSMDS. In each setting, the mean and standard deviation of the RMSE values of RMDS-AAP are computed over 1000 simulations. From Fig.~\ref{fig:noisy}, one can see that regardless of the noise strength, RMDS-AAP generally produces reconstruction with lower RMSE values, as well as smaller variations. 

\section{Proofs}

We first collect some useful results in Section \ref{sec:3.1}. Some of them are from the literature, and the proofs for the new ones are included in the Appendix. We then present the proofs for Theorem \ref{thm:main}, splitted into the proof for the initialization phase, and the proof for the iteration phase, in Section \ref{sec:3.2}, and Section \ref{sec:3.3}, respectively. The proof for Proposition \ref{prop} is presented in Section \ref{sec:3.4}. 

\subsection{Useful Lemmas}\label{sec:3.1}

\begin{lemma}[{\cite[Lemma 19]{Ding2020}}]\label{lem:L_infinity}
Let $L^{\star} = U^{\star}\Lambda^{\star}(U^{\star})^T$, and $L=\mathcal{H}_r^{+}\left(L^{\star}+E\right)$ for some perturbation matrix $E\in\mathbb{S}^{n\times n}$. Denote the eigen-decomposition of $L$ as $U\Lambda U^T$. Suppose the rank-$r$ SVD of $\left(U^{\star}\right)^TU$ is $A\Si B^T$. Set $G=AB^T$ and $\Delta=U-U^{\star}G$. If $\|E\|_{2} \leq\frac12\lambda_r^{\star}$, then
$$
\begin{aligned}
\left\|L-L^{\star}\right\|_{\infty} \leq & \|\Delta\|_{2, \infty}\lb\|U\|_{2, \infty}+\|U^{\star}\|_{2, \infty}\rb\|\Lambda\|_{2}+(3+4\kappa)\left\|U^{\star}\right\|_{2, \infty}^2\|E\|_{2}.
\end{aligned}
$$
\end{lemma}

\begin{lemma}[{\cite[Lemma 1]{Ding2020}}]\label{lem:perturb_gt} Under the same conditions of Lemma \ref{lem:L_infinity},
$$
\begin{aligned}
\left\|\Lambda^{\star} G-G \Lambda^{\star}\right\|_{2} & \leq\left(2+\frac{2 \lambda_1^{\star}}{\lambda_r^{\star}-\|E\|_{2}}\right)\|E\|_{2}, \\
\left\|\Lambda^{\star} H-G \Lambda^{\star}\right\|_{2} & \leq\left(2+\frac{\lambda_1^{\star}}{\lambda_r^{\star}-\|E\|_{2}}\right)\|E\|_{2}.
\end{aligned}
$$
\end{lemma}

\begin{lemma}[{\cite[Lemma 47]{Ma2019}}]\label{lem:rotation_diff}
Assume the same conditions of Lemma \ref{lem:L_infinity}. Denote $X=U(\Lambda)^{\frac12}$, and $X^{\star}=U^{\star}(\Lambda^{\star})^{\frac12}$. Suppose the rank-$r$ SVD of $\left(X^{\star}\right)^TX$ is $Y\widetilde{\Si} Z^T$. Set $R=YZ^T$. Then,
$$
\|G-R\|_2\leq 28\kappa^{\frac32}\cdot\frac{\|E\|_2}{\lambda_r^{\star}}.
$$
\end{lemma}

\begin{lemma}[{\cite[Lemma 4]{netrapalli2014non}}]\label{lem:S_op}
If $S\in\mathbb{S}^{n\times n}$ is $\alpha$-sparse, i.e., $S$ has no more than $\alpha n$ nonzero entries per row (and column), then $\ln S\rn_2\leq \alpha n\cdot\ln S\rn_{\infty}$.
\end{lemma}

\begin{lemma}\label{lem:A}
The operator $\mathcal{A}$ satisfies:
\begin{enumerate}
    \item  $\forall Z \in \mathbb{S}^{n\times n}$, $\| \mathcal{A}(Z) \|_{\infty}\leq4\| Z \|_{\infty}$; 
    \item $\forall Z \in \mathbb{S}^{n\times n}$ with $Z\bm{1}=0$, $\mathcal{B}(\mathcal{A}(Z))=Z$.
\end{enumerate}
\end{lemma}

The proof of Lemma \ref{lem:A} is deferred to Appendix \ref{sec:proof_A_infinity}.

\begin{lemma}\label{lem:S_infinity} If $\|\mathcal{A}(L^k) - D^{\star}\|_{\infty} \leq \xi^k$, and $S^{k} = \T_{\xi^{k}}(D-\mathcal{A}(L^k))$,
then $\text{supp}(S^k) \subseteq \text{supp}(S^{\star})$, and
$$\|S^k - S^{\star}\|_{\infty} \leq \|\mathcal{A}(L^k) - D^{\star}\|_{\infty}+\xi^k.$$
\end{lemma}

The proof of Lemma \ref{lem:S_infinity} is deferred to Appendix \ref{sec:proof_S_infinity}.

\begin{lemma}\label{lem:J}
The matrix $J=I_n-\frac1n \bm{1}\bm{1}^T$ satisfies:
\begin{enumerate}
    \item $\|J\|_2 = 1$,
    \item $JU^{\star} = U^{\star}$,
    \item $\|JZ\|_{2,\infty}\leq 2\|Z\|_{2,\infty},~\forall Z\in\mathbb{R}^{n\times r}$.
\end{enumerate}
\end{lemma}

The proof of Lemma \ref{lem:J} is deferred to Appendix \ref{sec:proof_J}.

\begin{lemma}[{\cite[Lemma 6]{accaltprj2019}}]\label{lem:T_diff} 
Suppose $L$ is a rank-$r$ positive semi-definite matrix with the eigen-decomposition $L=U\Lambda(U)^T$, where $U\in\mathbb{R}^{n\times r}$. Let $\mathcal{P}_T$ be the projection onto the tangent space of the manifold of symmetric positive semi-definite matrices of rank $r$ at $L$, as defined in \eqref{eq:projection}. Then,
$$
\|(\mathcal{I}-\mathcal{P}_T)(L^{\star})\|_2\leq \frac{\|L-L^{\star}\|_2^2}{\lambda_1^{\star}}.
$$
\end{lemma}

\begin{lemma}[{\cite[Lemma 8]{accaltprj2019}}]\label{lem:T_op} 
Under the same conditions of Lemma \ref{lem:T_diff},
$$
\|\mathcal{P}_T(Z)\|_2\leq \frac43\|Z\|_2,~\forall Z\in\mathbb{S}^{n\times n}.
$$
\end{lemma}

\subsection{Proof for the Initialization Phase}\label{sec:3.2}

Considering $L^0$ as the zero matrix, $S^{0} = \T_{\xi^{0}}(D-\mathcal{A}(L^0))$. Since
$$
\|\mathcal{A}(L^0)-D^{\star}\|_{\infty} = \|D^{\star}\|_{\infty} \leq \xi^0,
$$
by Lemma \ref{lem:S_infinity}, $\text{supp}(S^0)\subseteq \text{supp}(S^{\star})$, and
$$
\|S^0 - S^{\star}\|_{\infty}\leq \|D^{\star}\|_{\infty}+\xi^0 \leq 4\|D^{\star}\|_{\infty}\leq 16\|L^{\star}\|_{\infty}\leq 16\frac{\mu r}{n}\lambda_1^{\star}, 
$$
where the third inequality follows from $D^{\star}=\mathcal{A}(L^{\star})$ and Lemma \ref{lem:A}. Denoting $E^0:=\mathcal{B}(S^0-S^{\star})$,
\begin{equation}\label{eq:op_init}
\begin{aligned}
\|E^0\|_2=\|\mathcal{B}(S^0-S^{\star})\|_2
    =&\frac12 \|J(S^0-S^{\star})J\|_2 \\
    \leq &\frac12\|S^0-S^{\star}\|_2 \\
    \leq &\frac{\alpha n}{2}\|S^0 - S^{\star}\|_{\infty} \\
    \leq & 2(\alpha n)\|D^{\star}\|_{\infty}\leq 8(\alpha \mu r)\lambda_1^{\star},
\end{aligned}
\end{equation}
where the first inequality follows from Lemma \ref{lem:J}, and the second inequality follows from Lemma \ref{lem:S_op}. Therefore $\|E^0\|_2$ is no more than $\frac{\lambda_r^{\star}}{2}$ if $\alpha\leq \frac{1}{16}\cdot\frac{1}{\mu r\kappa}$. 

Suppose $L^1=\mathcal{H}_r^{+}(L^{\star}-E^0)$ has the eigen-decomposition $U^1\Lambda^1(U^1)^T$, where $U^1\in\mathbb{R}^{n\times r}$, and $\Lambda^1=\text{diag}(\lambda_1^1,\cdots,\lambda_r^1)$. Lemma \ref{lem:L_infinity} is applicable once we have the row-wise bound of $\Delta^1:=U^1-U^{\star}G^1$, where $G^1$ is the best rotation matrix between $U^1$ and $U^{\star}$ computed from the SVD of $H^1:=(U^{\star})^TU^1$. Consider $i=1,\cdots,n$.
\begin{align*}
    e_i^T\Delta^1 = & e_i^TU^1-e_i^TU^{\star}G^1 \\
    = & e_i^T(L^{\star}-E^0)U^1(\Lambda^1)^{-1}-e_i^TU^{\star}G^1 \\
    = & e_i^TU^{\star}\Lambda^{\star}\left[(U^{\star})^TU^1(\Lambda^1)^{-1}-(\Lambda^{\star})^{-1}G^1\right]-e_i^TE^0U^1(\Lambda^1)^{-1} \\
    = & \underbrace{e_i^TU^{\star}\Lambda^{\star}\left[(U^{\star})^TU^1(\Lambda^{\star})^{-1}-(\Lambda^{\star})^{-1}G^1\right]}_{T_1}\\
    & +\underbrace{e_i^TU^{\star}\Lambda^{\star}(U^{\star})^TU^1\left[(\Lambda^{1})^{-1}-(\Lambda^{\star})^{-1}\right]}_{T_2}-\underbrace{e_i^TE^0U^1(\Lambda^1)^{-1}}_{T_3}. 
\end{align*}

\noindent\textbf{Bounding $T_1$.} 
\begin{align*}
    \|T_1\|_2 = & \|e_i^TU^{\star}\left[\Lambda^{\star}(U^{\star})^TU^1-G^1\Lambda^{\star}\right](\Lambda^{\star})^{-1}\|_2 \\
    \leq & \|e_i^TU^{\star}\|_2\cdot\|\Lambda^{\star}H^1-G^1\Lambda^{\star}\|_2\cdot\|(\Lambda^{\star})^{-1}\|_2 \\
    \leq & \sqrt{\frac{\mu r}{n}}\cdot (2+2\kappa)\|E^0\|_2\cdot \frac{1}{\lambda_r^{\star}},
\end{align*}
where in the last inequality, Lemma \ref{lem:perturb_gt} is applied with the bound $\|E^0\|_2\leq \frac12\lambda_r^{\star}$.

\noindent\textbf{Bounding $T_2$.} Due to Weyl's inequality,
$$
|\lambda_j^1-\lambda_j^{\star}| \leq \|E^0\|_2,~j=1,\cdots,r.
$$
$$
\|(\Lambda^{1})^{-1}-(\Lambda^{\star})^{-1}\|_2=\max_{j=1,\cdots,r} \left|\frac{1}{\lambda_j^1}-\frac{1}{\lambda_j^{\star}}\right|=\max_{j=1,\cdots,r} \frac{|\lambda_j^1-\lambda_j^{\star}|}{\lambda_j^1\lambda_j^{\star}}\leq 2\frac{\|E^0\|_2}{(\lambda_r^{\star})^2},
$$
where $\lambda_r^1\geq \frac12\lambda_r^{\star}$ is used in the last inequality. Therefore,
\begin{align*}
    \|T_2\|_2 \leq & \|e_i^TU^{\star}\|_2\cdot\|\Lambda^{\star}\|_2\cdot
    \|(U^{\star})^TU^1\|_2\cdot
    \|(\Lambda^{1})^{-1}-(\Lambda^{\star})^{-1}\|_2
    \leq \sqrt{\frac{\mu r}{n}}\cdot2\kappa\frac{\|E^0\|_2}{\lambda_r^{\star}}.
\end{align*}

\noindent\textbf{Bounding $T_3$.} 
\begin{align*}
    \|T_3\|_2\leq\|e_i^TE^0U^1\|_2\cdot\|(\Lambda^1)^{-1}\|_2\leq  \|e_i^TE^0U^1\|_2\cdot\frac{2}{\lambda_r^{\star}},
\end{align*}
therefore we just need to bound $\|e_i^TE^0U^1\|_2$.
\begin{align*}
    \|e_i^TE^0U^1\|_2 \leq \|e_i^TE^0U^{\star}G^1\|_2+\|e_i^TE^0(U^1-U^{\star}G^1)\|_2 = \|e_i^TE^0U^{\star}\|_2+\|e_i^TE^0\Delta^1\|_2.
\end{align*}

For the first term,
\begin{align*}
    \|e_i^TE^0U^{\star}\|_2 = \|e_i^T\mathcal{B}(S^0-S^{\star})U^{\star}\|_2
    = & \frac12\|e_i^TJ(S^0-S^{\star})JU^{\star}\|_2 \\
    = & \frac12\|e_i^TJ(S^0-S^{\star})U^{\star}\|_2 \\
    \leq & \frac12\|J(S^0-S^{\star})U^{\star}\|_{2,\infty} \\
    \leq & \|(S^0-S^{\star})U^{\star}\|_{2,\infty} \\
    \leq & (\alpha n)\|S^0-S^{\star}\|_{\infty}\cdot\sqrt{\frac{\mu r}{n}}
    \leq 4(\alpha n)\|D^{\star}\|_{\infty}\sqrt{\frac{\mu r}{n}},
\end{align*}
where the third equality and the second inequality follows from Lemma \ref{lem:J}, the third inequality follows from triangular inequality and the row-wise bound of $U^{\star}$. 

The second term can be bounded similarly. Since 
$$
\bm{1}^T(L^{\star}-E^0)=\bm{1}^T(L^{\star}-\mathcal{B}(S^0-S^{\star})) = 0,
$$
$\bm{1}$ is in the null space of $L^1$. From $\bm{1}^T(U^1-U^{\star}G^1) = 0$, we can get $J\Delta^1 = \Delta^1$, and 
\begin{align*}
    \|e_i^TE^0\Delta^1\|_2 \leq & \|(S^0-S^{\star})\Delta^1\|_{2,\infty} \\
    \leq & (\alpha n)\|S^0-S^{\star}\|_{\infty}\cdot\|\Delta^1\|_{2,\infty} \leq 4(\alpha n)\|D^{\star}\|_{\infty}\|\Delta^1\|_{2,\infty}.
\end{align*}
Therefore,
\begin{align*}
    \|T_3\|_2\leq & 8(\alpha n)\frac{\|D^{\star}\|_{\infty}}{\lambda_r^{\star}}\sqrt{\frac{\mu r}{n}}+8(\alpha n)\frac{\|D^{\star}\|_{\infty}}{\lambda_r^{\star}}\|\Delta^1\|_{2,\infty} \\
    \leq & 8(\alpha n)\frac{\|D^{\star}\|_{\infty}}{\lambda_r^{\star}}\sqrt{\frac{\mu r}{n}}+32(\alpha \mu r\kappa)\|\Delta^1\|_{2,\infty} \\
    \leq & 8(\alpha n)\frac{\|D^{\star}\|_{\infty}}{\lambda_r^{\star}}\sqrt{\frac{\mu r}{n}}+\frac12\|\Delta^1\|_{2,\infty}
\end{align*}
where in the second inequality, the bound $\|D^{\star}\|_{\infty}\leq4\frac{\mu r}{n}\lambda_1^{\star}$ is used again, and the last inequality holds if $\alpha\leq \frac{1}{64}\cdot\frac{1}{\mu r\kappa}$.

Combining the bounds of $T_1$ to $T_3$, and taking the maximum with respect to $i$,
\begin{align*}
    \|\Delta^1\|_{2,\infty} \leq (2+2\kappa)\frac{\|E^0\|_2}{\lambda_r^{\star}}\sqrt{\frac{\mu r}{n}}+2\kappa\frac{\|E^0\|_2}{\lambda_r^{\star}}\sqrt{\frac{\mu r}{n}}+8(\alpha n)\frac{\|D^{\star}\|_{\infty}}{\lambda_r^{\star}}\sqrt{\frac{\mu r}{n}}+\frac12\|\Delta^1\|_{2,\infty},
\end{align*}
consequently,
\begin{equation}\label{eq:init_2_infty}
\begin{aligned}
    \|\Delta^1\|_{2,\infty} \leq &\left[12\kappa\frac{\|E^0\|_2}{\lambda_r^{\star}}+16(\alpha n)\frac{\|D^{\star}\|_{\infty}}{\lambda_r^{\star}}\right]\sqrt{\frac{\mu r}{n}} \\
    \leq & \left[24(\alpha \kappa n)\frac{\|D^{\star}\|_{\infty}}{\lambda_r^{\star}}+16(\alpha n)\frac{\|D^{\star}\|_{\infty}}{\lambda_r^{\star}}\right]\sqrt{\frac{\mu r}{n}} \\
    \leq & 40(\alpha\kappa n)\frac{\|D^{\star}\|_{\infty}}{\lambda_r^{\star}}\sqrt{\frac{\mu r}{n}} 
    \leq 160(\alpha\mu r\kappa^2)\sqrt{\frac{\mu r}{n}}\leq\sqrt{\frac{\mu r}{n}}
\end{aligned}
\end{equation}
where in the second inequality, the bound $\|E^0\|_2\leq 2(\alpha n)\|D^{\star}\|_{\infty}$ is used, and the last inequality holds if $\alpha\leq \frac{1}{160}\cdot\frac{1}{\mu r\kappa^2}$. As a result,
$$
\|e_i^TU^1\|_2 \leq \|e_i^TU^{\star}G^1\|_2+\|e_i^T\Delta^1\|_2 = \|e_i^TU^{\star}\|_2+\|e_i^T\Delta^1\|_2,
$$
therefore $\|U^1\|_{2,\infty}\leq 2\sqrt{\frac{\mu r}{n}}$. Also, under this bound of $\alpha$ we have
$$
\|E^0\|_2\leq 2(\alpha n)\|D^{\star}\|_{\infty} \leq 8(\alpha\mu r)\lambda_1^{\star}\leq \frac{1}{20}\lambda_r^{\star}.
$$
Now by Lemma \ref{lem:L_infinity},
\begin{equation}\label{eq:init_L_infty}
\begin{aligned}
\left\|L^1-L^{\star}\right\|_{\infty} \leq & \|\Delta^1\|_{2, \infty}\lb\|U^1\|_{2, \infty}+\|U^{\star}\|_{2, \infty}\rb\|\Lambda^1\|_{2}+(3+4\kappa)\left\|U^{\star}\right\|_{2, \infty}^2\|E^0\|_{2}\\
\leq & 40(\alpha\kappa n)\frac{\|D^{\star}\|_{\infty}}{\lambda_r^{\star}}\sqrt{\frac{\mu r}{n}}\cdot3\sqrt{\frac{\mu r}{n}}\cdot\frac{21}{20}\lambda_1^{\star}+7\kappa\frac{\mu r}{n}\cdot 2(\alpha n)\|D^{\star}\|_{\infty}\\
= & (126\alpha\mu r\kappa^2+14\alpha\mu r\kappa)\|D^{\star}\|_{\infty} \leq \frac{\|D^{\star}\|_{\infty}}{4}\gamma
\end{aligned}
\end{equation}
if $\alpha\leq \frac{1}{560}\cdot\frac{\gamma}{\mu r\kappa^2}$.

\subsection{Proof for the Iteration Phase}\label{sec:3.3}

For $k\geq 1$, our induction hypotheses are the following:
\begin{subequations}
\begin{align}
    \|E^{k-1}\|_2 \leq & 4(\alpha n)\|D^{\star}\|_{\infty}\gamma^{k-1}, \label{eq:op}\\
    \|\Delta^{k}\|_{2,\infty} \leq & 120(\alpha\kappa n)\frac{\|D^{\star}\|_{\infty}}{\lambda_r^{\star}}\sqrt{\frac{\mu r}{n}}\gamma^{k-1},\label{eq:2_infty} \\
    \|L^k-L^{\star}\|_{\infty} \leq & \frac{\|D^{\star}\|_{\infty}}{4}\gamma^k.\label{eq:L_infty}
\end{align}
\end{subequations}
One can see from \eqref{eq:op_init}, \eqref{eq:init_2_infty}, and \eqref{eq:init_L_infty} that the three bounds hold for $k=1$. For any $k\geq 1$,
$$
\|E^{k-1}\|_2\leq 16(\alpha\mu r)\lambda_1^{\star}\leq \frac{1}{20}\lambda_r^{\star}
$$
when $\alpha\leq \frac{1}{320}\cdot\frac{1}{\mu r\kappa}$; and
$$
\|\Delta^k\|_{2,\infty}\leq 480(\alpha\mu r\kappa^2)\sqrt{\frac{\mu r}{n}}\leq \sqrt{\frac{\mu r}{n}} 
$$
when $\alpha\leq \frac{1}{480}\cdot\frac{1}{\mu r\kappa^2}$ so that $\|U^k\|_{2,\infty}\leq 2\sqrt{\frac{\mu r}{n}}$. 

Now we prove for iteration $(k+1)$. With the bound 
$\|L^k-L^{\star}\|_{\infty} \leq \frac{\|D^{\star}\|_{\infty}}{4}\gamma^k$, by Lemma \ref{lem:A},
$$
\|\mathcal{A}(L^k)-D^{\star}\|_{\infty} \leq \|D^{\star}\|_{\infty}\gamma^k\leq \xi^k.
$$
Then by Lemma \ref{lem:S_infinity}, $\text{supp}(S^k)\subseteq \text{supp}(S^{\star})$, and
$$
\|S^k-S^{\star}\|_{\infty}\leq \|D^{\star}\|_{\infty}\gamma^k+\xi^k \leq (4\|D^{\star}\|_{\infty})\gamma^k.
$$
By Lemma \ref{lem:A}, $\mathcal{B}(\mathcal{A}(L^{\star}))=L^{\star}$, and
\begin{align*}
L^{k+1} = \mathcal{H}_r^{+}\mathcal{P}_{T^k}\mathcal{B}(D-S^k)
= &\mathcal{H}_r^{+}\mathcal{P}_{T^k}\mathcal{B}(D^{\star}+S^{\star}-S^k)\\
= &\mathcal{H}_r^{+}\mathcal{P}_{T^k}\mathcal{B}(\mathcal{A}(L^{\star})+S^{\star}-S^k)
= \mathcal{H}_r^{+}\mathcal{P}_{T^k}(L^{\star}+\mathcal{B}(S^{\star}-S^k)). 
\end{align*}
Denoting
\begin{align*}
L^{k+1} =\mathcal{H}_r^{+}\mathcal{P}_{T^k}(L^{\star}+\mathcal{B}(S^{\star}-S^k))
= &\mathcal{H}_r^{+}(L^{\star}-((\mathcal{I}-\mathcal{P}_{T^k})L^{\star}+\mathcal{P}_{T^k}\mathcal{B}(S^k-S^{\star})):=\mathcal{H}_r^{+}(L^{\star}-E^{k}),
\end{align*}
and the eigen-decomposition of $L^{k+1}$ as $U^{k+1}\Lambda^{k+1}(U^{k+1})^T$.
\begin{align*}
    \|E^{k}\|_2 = &\|(\mathcal{I}-\mathcal{P}_{T^k})L^{\star}+\mathcal{P}_{T^k}\mathcal{B}(S^k-S^{\star})\|_2 \\
    \leq &\frac{\|L^k-L^{\star}\|_2^2}{\lambda_1^{\star}}+\frac43\|\mathcal{B}(S^k-S^{\star})\|_2 \\
    \leq &\frac{\|E^{k-1}\|_2}{\lambda_1^{\star}}\|E^{k-1}\|_2+\frac23\|S^k-S^{\star}\|_2 \\
    \leq &4(\alpha n)\frac{\|D^{\star}\|_{\infty}}{\lambda_1^{\star}}\gamma^{k-1}\cdot4(\alpha n)\|D^{\star}\|_{\infty}\gamma^{k-1}+\frac83(\alpha n)\|D^{\star}\|_{\infty}\gamma^k\\
    \leq & 16(\alpha \mu r)\gamma^{k-1}\cdot4(\alpha n)\|D^{\star}\|_{\infty}\gamma^{k-1}+\frac83(\alpha n)\|D^{\star}\|_{\infty}\gamma^k\\
    \leq & \frac{\gamma}{3}\cdot4(\alpha n)\|D^{\star}\|_{\infty}\gamma^{k-1}+\frac83(\alpha n)\|D^{\star}\|_{\infty}\gamma^k
    = 4(\alpha n)\|D^{\star}\|_{\infty}\gamma^k,
\end{align*}
where the first inequality follows from Lemma \ref{lem:T_diff} and Lemma \ref{lem:T_op}, the second inequality is due to Lemma \ref{lem:J} again, the third inequality follows from the induction hypothesis \eqref{eq:op},  
and the fifth inequality holds if $\alpha\leq \frac{1}{48}\cdot\frac{\gamma}{\mu r}$. Then we proceed as in the base case to get
\begin{align*}
    e_i^T\Delta^{k+1} = & \underbrace{e_i^TU^{\star}\Lambda^{\star}\left[(U^{\star})^TU^{k+1}(\Lambda^{\star})^{-1}-(\Lambda^{\star})^{-1}G^{k+1}\right]}_{T_1}\\
    & +\underbrace{e_i^TU^{\star}\Lambda^{\star}(U^{\star})^TU^{k+1}\left[(\Lambda^{k+1})^{-1}-(\Lambda^{\star})^{-1}\right]}_{T_2}-\underbrace{e_i^TE^kU^{k+1}(\Lambda^{k+1})^{-1}}_{T_3}, 
\end{align*}
and the first two terms can be bounded similarly, i.e.,
\begin{align*}
    \|T_1\|_2 \leq \sqrt{\frac{\mu r}{n}} \cdot (2+2\kappa)\frac{\|E^k\|_2}{\lambda_r^{\star}},~
    \|T_2\|_2 \leq \sqrt{\frac{\mu r}{n}} \cdot 2\kappa\frac{\|E^k\|_2}{\lambda_r^{\star}}.
\end{align*}

\noindent\textbf{Bounding $T_3$.} 
\begin{align*}
    \|T_3\|_2\leq\|e_i^TE^kU^{k+1}\|_2\cdot\|(\Lambda^{k+1})^{-1}\|_2\leq\|e_i^TE^k\|_2\cdot\frac{2}{\lambda_r^{\star}},
\end{align*}
while
\begin{align*}
    \|e_i^TE^k\|_2\leq\underbrace{\|e_i^T(\mathcal{I}-\mathcal{P}_{T^k})L^{\star}\|_2}_{B_1}+\underbrace{\|e_i^T\mathcal{P}_{T^k}\mathcal{B}(S^k-S^{\star})\|_2}_{B_2}.
\end{align*}

For the first term, 
\begin{align*}
    (\mathcal{I}-\mathcal{P}_{T^k})L^{\star} = & (I_n-U^{k}(U^{k})^T)L^{\star}(I_n-U^{k}(U^{k})^T) \\
    = & (U^{k}(U^{k})^T-U^{\star}(U^{\star})^T)(-L^{\star})(I_n-U^{k}(U^{k})^T) \\
    = & (U^{k}(U^{k})^T-U^{\star}(U^{\star})^T)(L^k-L^{\star})(I_n-U^{k}(U^{k})^T)
\end{align*}
where the last equality follows from the fact that $L^k(I_n-U^{k}(U^{k})^T)=0$. Therefore,
\begin{align*}
    B_1 = & \|e_i^T(U^{k}(U^{k})^T-U^{\star}(U^{\star})^T)(L^k-L^{\star})(I_n-U^{k}(U^{k})^T)\|_2 \\
    \leq & \|e_i^T(U^{k}(U^{k})^T-U^{\star}(U^{\star})^T)\|_2\cdot\|L^k-L^{\star}\|_2 \\
    \leq & (\|e_i^TU^k\|_2+\|e_i^TU^{\star}\|_2)\|L^k-L^{\star}\|_2
    \leq 3\sqrt{\frac{\mu r}{n}}\cdot\|E^{k-1}\|_2.
\end{align*}
For the second term,
\begin{align*}
    B_2 = & \|e_i^T\mathcal{P}_{T^k}\mathcal{B}(S^k-S^{\star})\|_2 \\
    \leq & \underbrace{\|e_i^TU^k(U^k)^T\mathcal{B}(S^k-S^{\star})(I_n-U^k(U^k)^T)\|_2}_{B_{21}}+\underbrace{\|e_i^T\mathcal{B}(S^k-S^{\star})U^k(U^k)^T\|_2}_{B_{22}}. 
\end{align*}
\begin{align*}
    B_{21} \leq \|e_i^TU^k\|_2\cdot\|\mathcal{B}(S^k-S^{\star})\|_2
    \leq 2\sqrt{\frac{\mu r}{n}}\cdot\frac{\alpha n}{2}\|S^k-S^{\star}\|_{\infty}
    \leq \sqrt{\frac{\mu r}{n}}\cdot 4(\alpha n)\|D^{\star}\|_{\infty}\gamma^k,
\end{align*}
\begin{align*}
    B_{22} \leq \|e_i^T\mathcal{B}(S^k-S^{\star})U^k\|_2
    \leq &\frac12\|J(S^k-S^{\star})JU^k\|_{2,\infty} \\
    = &\frac12\|J(S^k-S^{\star})U^k\|_{2,\infty} \\
    \leq & \|(S^k-S^{\star})U^k\|_{2,\infty}\leq 4(\alpha n)\|D^{\star}\|_{\infty}\gamma^k\cdot2\sqrt{\frac{\mu r}{n}}.
\end{align*}

Putting together,

\begin{align*}
    \|T_3\|_2\leq 6\frac{\|E^{k-1}\|_2}{\lambda_r^{\star}}\sqrt{\frac{\mu r}{n}}+24(\alpha n)\frac{\|D^{\star}\|_{\infty}}{\lambda_r^{\star}}\sqrt{\frac{\mu r}{n}}\gamma^k.
\end{align*}

Combining the bounds of $T_1$ to $T_3$, and taking the maximum with respect to $i$,

\begin{align*}
    \|\Delta^{k+1}\|_{2,\infty}\leq & 6\kappa\frac{\|E^{k}\|_2}{\lambda_r^{\star}}\sqrt{\frac{\mu r}{n}}+6\frac{\|E^{k-1}\|_2}{\lambda_r^{\star}}\sqrt{\frac{\mu r}{n}}+24(\alpha n)\frac{\|D^{\star}\|_{\infty}}{\lambda_r^{\star}}\sqrt{\frac{\mu r}{n}}\gamma^k\\
    \leq & \left[24(\alpha\kappa n)\frac{\|D^{\star}\|_{\infty}}{\lambda_r^{\star}}\gamma^k+24(\alpha n)\frac{\|D^{\star}\|_{\infty}}{\lambda_r^{\star}}\gamma^{k-1}+24(\alpha n)\frac{\|D^{\star}\|_{\infty}}{\lambda_r^{\star}}\gamma^{k}\right]\sqrt{\frac{\mu r}{n}} \\
    \leq & \left[24(\alpha\kappa n)\frac{\|D^{\star}\|_{\infty}}{\lambda_r^{\star}}\gamma^k+72(\alpha n)\frac{\|D^{\star}\|_{\infty}}{\lambda_r^{\star}}\gamma^{k}+24(\alpha n)\frac{\|D^{\star}\|_{\infty}}{\lambda_r^{\star}}\gamma^{k}\right]\sqrt{\frac{\mu r}{n}} \\
    \leq & 120(\alpha\kappa n)\frac{\|D^{\star}\|_{\infty}}{\lambda_r^{\star}}\sqrt{\frac{\mu r}{n}}\gamma^k,
\end{align*}
where the third inequality holds if $\gamma\geq \frac{1}{3}$. By Lemma \ref{lem:L_infinity} again,

\begin{align*}
\left\|L^{k+1}-L^{\star}\right\|_{\infty} \leq & \|\Delta^{k+1}\|_{2, \infty}\lb\|U^{k+1}\|_{2, \infty}+\|U^{\star}\|_{2, \infty}\rb\|\Lambda^{k+1}\|_{2}+(3+4\kappa)\left\|U^{\star}\right\|_{2, \infty}^2\|E^k\|_{2}\\
\leq & 120(\alpha\kappa n)\frac{\|D^{\star}\|_{\infty}}{\lambda_r^{\star}}\sqrt{\frac{\mu r}{n}}\gamma^k\cdot3\sqrt{\frac{\mu r}{n}}\cdot\frac{21}{20}\lambda_1^{\star}+7\kappa\frac{\mu r}{n}\cdot (4\alpha n)\|D^{\star}\|_{\infty}\gamma^k \\
\leq &406(\alpha \mu r\kappa^2) \|D^{\star}\|_{\infty}\gamma^k\leq\frac{\|D^{\star}\|_{\infty}}{4}\gamma^{k+1} 
\end{align*}
if $\alpha\leq \frac{1}{1624}\cdot\frac{\gamma}{\mu r\kappa^2}$.

\subsection{Proof for Proposition \ref{prop}}\label{sec:3.4}

For $k\geq 0$,
\begin{align*}
    &\|X^{k+1}-X^{\star}R^{k+1}\|_{2,\infty} \\
    = & \|U^{k+1}(\Lambda^{k+1})^{\frac12}-U^{\star}(\Lambda^{\star})^{\frac12}R^{k+1}\|_{2,\infty} \\
    \leq & \underbrace{\|U^{k+1}(\Lambda^{k+1})^{\frac12}-U^{\star}G^{k+1}(\Lambda^{k+1})^{\frac12}\|_{2,\infty}}_{I_1}+\underbrace{\|U^{\star}G^{k+1}(\Lambda^{k+1})^{\frac12}-U^{\star}G^{k+1}(\Lambda^{\star})^{\frac12}\|_{2,\infty}}_{I_2}\\
    &+\underbrace{\|U^{\star}G^{k+1}(\Lambda^{\star})^{\frac12}-U^{\star}(\Lambda^{\star})^{\frac12}G^{k+1}\|_{2,\infty}}_{I_3}+\underbrace{\|U^{\star}(\Lambda^{\star})^{\frac12}G^{k+1}-U^{\star}(\Lambda^{\star})^{\frac12}R^{k+1}\|_{2,\infty}}_{I_4}.
\end{align*}

\noindent{\textbf{Bounding $I_1$.}}
\begin{align*}
    I_1 = &\max_i\|e_i^T[U^{k+1}(\Lambda^{k+1})^{\frac12}-U^{\star}G^{k+1}(\Lambda^{k+1})^{\frac12}]\|_2 \\
    \leq &\|\Delta^{k+1}\|_{2,\infty}\cdot\|(\Lambda^{k+1})^{\frac12}\|_2 \\
    \leq & 120(\alpha\kappa n)\frac{\|D^{\star}\|_{2,\infty}}{\lambda_r^{\star}}\sqrt{\frac{\mu r}{n}}\gamma^k\cdot\left(\frac{21}{20}\lambda_1^{\star}\right)^{\frac12} \leq 492(\alpha\mu r\kappa^2 )\sqrt{\frac{\mu r}{n}}(\lambda_1^{\star})^{\frac12}\cdot\gamma^k.
\end{align*}

\noindent{\textbf{Bounding $I_2$.}}
\begin{align*}
    I_2 = \max_i\|e_i^TU^{\star}G^{k+1}[(\Lambda^{k+1})^{\frac12}-(\Lambda^{\star})^{\frac12}]\|_2
    \leq \sqrt{\frac{\mu r}{n}}\cdot\|(\Lambda^{k+1})^{\frac12}-(\Lambda^{\star})^{\frac12}\|_2.
\end{align*}
Since
\begin{align*}
\|(\Lambda^{k+1})^{\frac12}-(\Lambda^{\star})^{\frac12}\|_2=\max_{j=1,\cdots,r} \left|(\lambda_j^{k+1})^{\frac12}-(\lambda_j^{\star})^{\frac12}\right|=\max_{j=1,\cdots,r} \frac{|\lambda_j^{k+1}-\lambda_j^{\star}|}{(\lambda_j^{k+1})^{\frac12}+(\lambda_j^{\star})^{\frac12}}\leq \frac{\|E^k\|_2}{(\lambda_r^{\star})^{\frac12}},
\end{align*}
$$
I_2\leq 4(\alpha n)\frac{\|D^{\star}\|_{\infty}}{(\lambda_r^{\star})^{\frac12}}\sqrt{\frac{\mu r}{n}}\cdot\gamma^k
\leq 16(\alpha\mu r\kappa^{\frac12})\sqrt{\frac{\mu r}{n}}(\lambda_1^{\star})^{\frac12}\cdot\gamma^k.
$$

\noindent{\textbf{Bounding $I_3$.}} 
\begin{align*}
    I_3 = \max_i\|e_i^TU^{\star}[G^{k+1}(\Lambda^{\star})^{\frac12}-(\Lambda^{\star})^{\frac12}G^{k+1}]\|_2
    \leq \sqrt{\frac{\mu r}{n}}\cdot\|G^{k+1}(\Lambda^{\star})^{\frac12}-(\Lambda^{\star})^{\frac12}G^{k+1}\|_2.
\end{align*}
\begin{align*}
    \|G^{k+1}(\Lambda^{\star})^{\frac12}-(\Lambda^{\star})^{\frac12}G^{k+1}\|_2 = & \|G^{k+1}(\Lambda^{\star})^{\frac12}-(\Lambda^{\star})^{\frac12}G^{k+1}\|_2 \\
    = & \|(\Lambda^{\star})^{\frac12}-(G^{k+1})^T(\Lambda^{\star})^{\frac12}G^{k+1}\|_2 \\
    \leq & \frac{1}{2(\lambda_r^{\star})^{\frac12}} \|\Lambda^{\star}-(G^{k+1})^T\Lambda^{\star}G^{k+1}\|_2 \\
    = &\frac{1}{2(\lambda_r^{\star})^{\frac12}} \|\Lambda^{\star}G^{k+1}-G^{k+1}\Lambda^{\star}\|_2\\
    \leq &\frac{1}{2(\lambda_r^{\star})^{\frac12}}\cdot(2+4\kappa)\|E^k\|_2\leq 3\kappa\frac{\|E^k\|_2}{(\lambda_r^{\star})^{\frac12}},
\end{align*}
where the first inequality follows from \cite[Lemma 2.1]{Schmitt1992}, since $(\Lambda^{\star})^{\frac12}$ and $(G^{k+1})^T(\Lambda^{\star})^{\frac12}G^{k+1}$ are the matrix square root of $\Lambda^{\star}$ and $(G^{k+1})^T\Lambda^{\star}G^{k+1}$, respectively; and the second inequality follows from Lemma \ref{lem:perturb_gt}. Therefore,
$$
I_3\leq 12(\alpha\kappa n)\frac{\|D^{\star}\|_{\infty}}{(\lambda_r^{\star})^{\frac12}}\sqrt{\frac{\mu r}{n}}\cdot\gamma^k
\leq 48(\alpha\mu r\kappa^{\frac32})\sqrt{\frac{\mu r}{n}}(\lambda_1^{\star})^{\frac12}\cdot\gamma^k.
$$

\noindent{\textbf{Bounding $I_4$.}} 
\begin{align*}
    I_4 = &\max_i\|e_i^T[U^{\star}(\Lambda^{\star})^{\frac12}G^{k+1}-U^{\star}(\Lambda^{\star})^{\frac12}R^{k+1}]\|_2 \\
    \leq & \sqrt{\frac{\mu r}{n}}\cdot(\lambda_1^{\star})^{\frac12}\cdot\|G^{k+1}-R^{k+1}\|_2\\
    \leq & \sqrt{\frac{\mu r}{n}}\cdot(\lambda_1^{\star})^{\frac12}\cdot28\kappa^{\frac32}\frac{\|E^k\|_2}{\lambda_r^{\star}} \leq 112(\alpha \kappa^2n)\frac{\|D^{\star}\|_{\infty}}{(\lambda_r^{\star})^{\frac12}}\sqrt{\frac{\mu r}{n}}\cdot\gamma^k\leq 448(\alpha\mu r\kappa^{\frac52})\sqrt{\frac{\mu r}{n}}(\lambda_1^{\star})^{\frac12}\cdot\gamma^k.
\end{align*}
where the second inequality is due to Lemma \ref{lem:rotation_diff}.

Combining the bound of $I_1$ to $I_4$, 
$$
\|X^{k+1}-X^{\star}R^{k+1}\|_{2,\infty} \leq 1004(\alpha\mu r\kappa^{\frac52})\sqrt{\frac{\mu r}{n}}(\lambda_1^{\star})^{\frac12}\cdot\gamma^k \leq \sqrt{\frac{\mu r\kappa\lambda_1^{\star}}{n}}\gamma^{k+1},
$$
where the last inequality follows from $\alpha\leq \frac{1}{1624}\cdot\frac{\gamma}{\mu r\kappa^2}$.

\section*{Conclusion}

In this paper, we propose an alternating projection based algorithm that is further accelerated by the tangent space projection technique, for the RMDS problem. Under standard assumptions in the RPCA literature, we establish linear convergence of the proposed algorithm when the outliers are sparse enough. We also numerically verified the performances of the proposed algorithm, with comparisons to other state-of-the-art solvers for RMDS.

To adapt to more realistic settings, it is of interest to incorporate noise in the convergence proof to handle the case that the pairwise distances are corrupted by sub-Gaussian noise, in addition to sparse outliers. 

\bibliographystyle{siam}
\bibliography{ref}

\appendix

\section{Implementation Details of the Computation of $L^{k+1}$}\label{sec:implementation}

First consider $\mathcal{B}(D-S^k)=-\frac12J(D-S^k)J$. The computation of $J(D-S^k)$ can be achieved by subtracting from each row of $(D-S^k)$ by its averaged row vector. This step costs about $2n^2$ flops. $(J(D-S^k))J$ can be computed in a similar fashion, by subtracting from each column of $J(D-S^k)$ by its averaged column vector. Therefore, the computation cost of $\mathcal{B}(D-S^k)$ is about $5n^2$ flops.

Denote $B:=\mathcal{B}(D-S^k)$. Note that $(I-U^k(U^k)^T)BU^k=BU^k-U^k(U^k)^TBU^k$, and the computation of $U^k(U^k)^TBU^k$ requires about $2n^2r+4nr^2$ flops. Therefore the total cost to form $(I-U^k(U^k)^T) B U^k$ is about $nr+2n^2r+4nr^2$ flops. Let $(I-U^k(U^k)^T)B U^k = QR$ be the QR decomposition, whose cost is $O(nr^2)$ flops. Then,
\begin{align*}
    \P_{T^k}(B) &= U^k (U^k)^T B + B U^k (U^k)^T - U^k (U^k)^T B U^k (U^k)^T \\
&= U^k (U^k)^T B (I - U^k (U^k)^T) + (I - U^k (U^k)^T) B U^k (U^k)^T \\
&\quad + U^k (U^k)^T B U^k (U^k)^T \\
&= U^k R^T Q^T + Q R (U^k)^T + U^k (U^k)^T B U^k (U^k)^T \\
&= [U^k~Q] \begin{bmatrix} (U^k)^T B U^k & R^T \\ R & 0 \end{bmatrix} \begin{bmatrix} (U^k)^T \\ Q^T \end{bmatrix} \\
&:= [U^k~Q] M ([U^k~Q])^T.
\end{align*}

Note that $M\in\mathbb{S}^{2r\times 2r}$, the eigen-decomposition of $M$, denoted as $U \Lambda U^T$ can be computed using $O(r^3)$ flops. We may as well require that the diagonal entries of $\Lambda$ are in the descending order. Finally, since the columns of $Q$ is orthogonal to the columns of $U^k$, $[U^k~Q]$ is an orthogonal matrix, and we can get the eigen-decomposition of $L^{k+1}$ by computing
\begin{equation}
U^{k+1} = \begin{bmatrix} U^k & Q \end{bmatrix} U_{(:,1:r)}. 
\end{equation}
The cost of this step is about $4nr^2$. In summary, the computational cost of $\mathcal{H}_r^{+}\P_{T^k}\mathcal{B}(D-S^k)$ is about $5n^2+2n^2r+O(nr^2)$ flops.

\section{Proofs for Some Useful Lemmas}
\subsection{Proof for Lemma \ref{lem:A}}\label{sec:proof_A_infinity}

By the definition of operator $\mathcal{A}$,
\begin{align*}
    \mathcal{A}(Z)=\text{diag}(Z)\bm{1}^T+\bm{1}\text{diag}(Z)^T-2Z.
\end{align*} 
Therefore $[\mathcal{A}(Z)]_{ij}=Z_{ii}+Z_{jj}-2Z_{ij}$,
\begin{align*}
    \|\mathcal{A}(Z)\|_{\infty}
    \leq\|Z \|_{\infty}+\|Z \|_{\infty}+2\| Z \|_{\infty}
    =4\|Z \|_{\infty}.
\end{align*}
\begin{align*}
\mathcal{B}(\mathcal{A}(Z)) = &-\frac12 J(\text{diag}(Z)\bm{1}^T+\bm{1}\text{diag}(Z)^T-2Z)J \\
= & -\frac12J\cdot\text{diag}(Z)\bm{1}^T(I_n-\frac1n\bm{1}\bm{1}^T)-\frac12(I_n-\frac1n\bm{1}\bm{1}^T)\cdot\bm{1}\text{diag}(Z)^TJ+JZJ = JZJ,
\end{align*}
and if $Z\bm{1}=0$,
\begin{align*}
    JZJ = JZ(I_n-\frac1n\bm{1}\bm{1}^T) = JZ = (I_n-\frac1n\bm{1}\bm{1}^T)Z= Z.
\end{align*}

\subsection{Proof for Lemma \ref{lem:S_infinity}}\label{sec:proof_S_infinity}

Denote $D^k=\mathcal{A}(L^k)$.
\begin{align*}
 S^k = T_{\xi^k}(D-\mathcal{A}(L^k))=T_{\xi^k}(S^{\star}+D^{\star}-D^k).
\end{align*}
When $S_{ij}^{\star}=0$, since $\|D^k-D^{\star}\|_{\infty}\leq \xi^k$,
\begin{align*}
     S_{ij}^k=T_{\xi^k}(D_{ij}^{\star}-D_{ij}^k)= 0,
\end{align*}
and it follows that $\text{supp}(S^k)\subseteq \text{supp}(S^{\star})$. Next, there are two cases to consider.
\begin{enumerate}
    \item $(i,j) \in \text{supp}(S^{\star}) \setminus \text{supp}(S^k)$. In this case, $|S_{ij}^{\star} + D_{ij}^{\star} - D_{ij}^k|\leq\xi^k$,
    \begin{align*}
     | S_{ij}^k - S_{ij}^{\star} | = | S_{ij}^{\star} | \leq | D_{ij}^k - D_{ij}^{\star} | + \xi^k.
     \end{align*}
    \item $(i,j) \in \text{supp}(S^k)$. In this case, $|S_{ij}^{\star} + D_{ij}^{\star} - D_{ij}^k|>\xi^k$,
    \begin{align*}
        S_{ij}^k =T_{\xi^k}(S_{ij}^{\star}+D_{ij}^{\star}-D_{ij}^k)=S_{ij}^{\star}+D_{ij}^{\star}-D_{ij}^k,
    \end{align*}
    and we get the bound
     \begin{align*}
         | S_{ij}^k - S_{ij}^{\star} | = | D_{ij}^k - D_{ij}^{\star} | \leq \xi^k.
     \end{align*}    
\end{enumerate}

\subsection{Proof for Lemma \ref{lem:J}}\label{sec:proof_J}

It is easy to see that $\|J\|_2\leq 1$, and any unit vector orthogonal to $\bm{1}$ reaches the upper bound. For the second property, recall that $L^{\star}=U^{\star}\Lambda^{\star}(U^{\star})^T$, and $L^{\star}\bm{1}=0$, therefore $\bm{1}$ is in the null space of $L^{\star}$, and is orthogonal to each column of $U^{\star}$. Hence,
$$
\bm{1}^TU^{\star}=0,~JU^{\star}=U^{\star}-\frac1n \bm{1}\bm{1}^TU^{\star}=U^{\star}.
$$
Finally, denote the $i$-th row of $Z$ as $z_i$,
$$
\|e_i^T(JZ)\|_2 = \left\|z_i-\frac{z_1+\cdots+z_n}{n}\right\|_2\leq 2\|Z\|_{2,\infty}. 
$$

\end{document}